\documentclass[conference]{IEEEtran}
\IEEEoverridecommandlockouts
\usepackage{cite}
\usepackage{amsmath,amssymb,amsfonts}
\usepackage{algorithmic}
\usepackage{graphicx}
\usepackage{textcomp}
\usepackage{xcolor}
\usepackage{float}
\usepackage{algorithm}
\usepackage{algorithmic}
\usepackage{xcolor}
\usepackage{comment}
\usepackage{subcaption}
\usepackage{tikz}
\usetikzlibrary{shapes.geometric, arrows.meta, positioning, fit, backgrounds, calc, shadows}
\usepackage{amsmath}
\usepackage{amssymb}
\usepackage{caption} 
\usepackage{booktabs}

\def\BibTeX{{\rm B\kern-.05em{\sc i\kern-.025em b}\kern-.08em
    T\kern-.1667em\lower.7ex\hbox{E}\kern-.125emX}}
\begin{document}

\title{Reinforcement Learning for Heterogeneous Sensor Selection in Maritime Surveillance\\
}
\author{%
\IEEEauthorblockN{%
\IEEEauthorrefmark{1}Andrei Starodubov,
\IEEEauthorrefmark{2}Yaqub Aris Prabowo,
\IEEEauthorrefmark{1}Andreas Hadjipieris,
\IEEEauthorrefmark{2}Roberto Galeazzi,
\IEEEauthorrefmark{1}Ioannis Kyriakides,
}
\IEEEauthorblockA{
andrei.starodubov@cmmi.blue,
yaqpr@dtu.dk,
andreas.hadjipieris@cmmi.blue,
roga@dtu.dk,
ioannis.kyriakides@cmmi.blue
}
\IEEEauthorblockA{\IEEEauthorrefmark{1}%
Cyprus Marine and Maritime Institute, Larnaca, Cyprus\
}
\IEEEauthorblockA{\IEEEauthorrefmark{2}%
Technical University of Denmark, Lyngby, Denmark\
}
}

\maketitle

\begin{abstract}
This paper presents an information-gain-guided reinforcement-learning sensor-selection framework for single-vessel tracking in heterogeneous maritime sensor networks. The proposed approach is motivated by information-theoretic sensor management: instead of activating all sensors or repeatedly performing computationally expensive online expected-information-gain evaluation, a learned policy selects one tracking-relevant sensor at each decision epoch. A Bayesian sequential Monte Carlo tracker estimates the vessel state from noisy measurements and provides a belief representation for scheduling under nonlinear and non-Gaussian conditions. A Proximal Policy Optimization agent selects one of five sensors in a georeferenced simulation of the CMMI Smart Marina testbed at Ayia Napa Marina, Cyprus. The policy is trained on the testbed’s actual five-sensor configuration. The agent observes belief-state, detection-history, coverage, sensor-geometry, and realized-information-gain features. The reward is defined as a realized-information-gain term gated by an observability mask. Final-test simulations compare the proposed framework with random single-sensor selection, always-on sensing using all sensors simultaneously, and the expected-information-gain sensor-selection baseline proposed in our previous work. Results show that the learned policy achieves tracking performance close to always-on sensing while activating only one sensor per decision time step and avoiding the computationally expensive online entropy search required by expected-information-gain selection. Additional zero-shot evaluation without retraining on ten moderately perturbed versions of actual layout configuration showed broadly stable tracking, with any increase in positional tracking error remaining below 1 meter across all perturbations.
\end{abstract}


\section{Introduction}

Persistent surveillance of small vessels in coastal ports and marinas is a critical component of maritime security. Non-cooperative craft in confined near-shore environments are difficult to monitor continuously because they may leave the field of view (FOV) of any single sensor due to various reasons~\cite{chang2003vessel}. Fixed networks of cameras and LiDARs can provide complementary coverage around a marina and support track continuity in such scenarios~\cite{lyu2022sea}. The resulting sensor-management problem has two coupled aspects: estimating the vessel state from heterogeneous measurements, and deciding which sensing asset should be used at each time step.

Bayesian sequential Monte Carlo (BSMC), implemented here as a particle filter (PF), provides a natural estimation framework for this setting because it represents the full posterior belief over the target state and can accommodate nonlinear dynamics, non-Gaussian uncertainty, and heterogeneous sensor likelihoods~\cite{arulampalam2002tutorial}. However, the PF itself is a model-based Bayesian estimator rather than a learned decision maker. Conversely, reinforcement learning (RL) provides a mechanism for learning sequential sensing decisions, but it requires a compact and meaningful representation of the tracking state and an objective that reflects sensing utility.

Information gain provides a principled connection between these two components. Since the PF maintains an approximation of the posterior belief distribution, sensor utility can be expressed through the reduction in posterior uncertainty induced by a measurement. This motivates information-theoretic sensor selection~\cite{kyriakides2021agile}, where candidate sensing actions are evaluated by their expected reduction in belief entropy. Such a criterion is directly defined on the Bayesian belief rather than only on a point estimate, making it well suited to uncertainty-aware sensor scheduling.

In our previous work, this principle was implemented through an expected information-gain sensor-selection (EIG-SS) method for camera--LiDAR maritime tracking~\cite{starodubov2026adaptive}. EIG-SS selected the sensing modality that produced the largest expected entropy reduction and therefore provided a principled adaptive sensing baseline. However, online expected-information-gain evaluation requires counterfactual PF updates for candidate sensors, whose computational cost increases with their numbers.

This paper addresses this limitation by replacing explicit online entropy search with a learned information-guided selection policy. We propose information-gain guided reinforcement-learning sensor selection (IG-RLSS), a proximal policy optimization (PPO)-based framework~\cite{schulman2017proximal} that shifts most of the decision-making cost to offline training. At inference time, the trained policy selects one sensor per decision time step from a fixed camera--LiDAR network, while the PF continues to provide the tracking belief. The policy is trained from PF-belief quantities, using realized posterior entropy reduction as an information-gain signal, so the learning objective is driven by belief-state uncertainty and sensor-use utility rather than by an explicit ground-truth tracking-error penalty.

To the best of our knowledge, BSMC/PF belief representations have been combined with reinforcement learning primarily in generic partially observable Markov Decision Processes (POMDP), robot-localization, and active-localization settings~\cite{ma2020dpfrl,jonschkowski2018differentiable,fischer2020ipft}, whereas RL-based target tracking and sensor scheduling have generally relied on non-SMC belief representations, mobile-agent control formulations, or proxy scheduling objectives~\cite{zheng2023novel,hoffmann2020sensorpath,qu2022improved,jeong2020learning,yang2023policy}. In contrast, this work uses a classical BSMC tracker as the belief source for a PPO-based sensor-selection policy in a fixed heterogeneous maritime surveillance network. The resulting scheduler can be interpreted as a belief-based, amortized approximation to information-driven sensing: it avoids repeated online counterfactual PF entropy searches over candidate sensors while being evaluated against an explicit EIG-SS baseline.

\section{Methodology}
\label{sec:method}

\begin{figure}[b]
\centering
\includegraphics[width=\columnwidth]{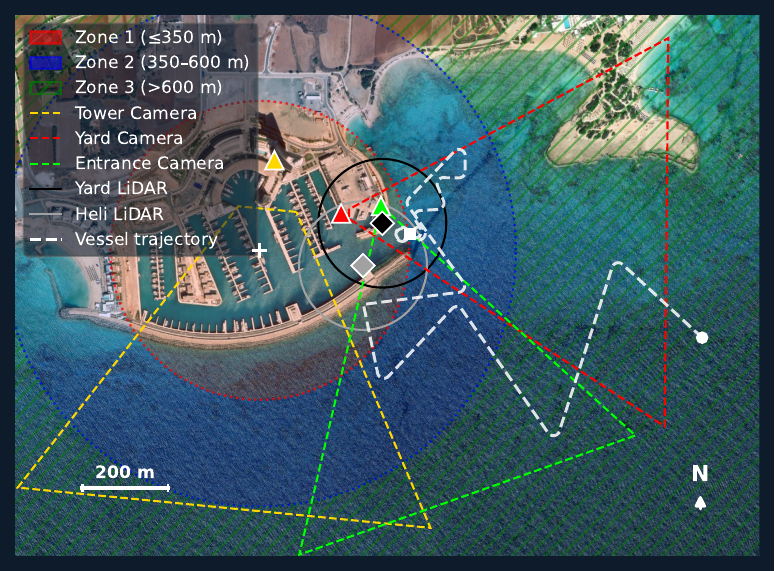}
\caption{Simulated marina environment showing sensor locations, fields of view, spatial evaluation zones, and an example vessel route.}
\label{fig:map}
\end{figure}

The simulated environment represents a georeferenced coastal-marina surveillance area of approximately $3.0~\mathrm{km}^2$ in a local projected coordinate frame, as shown in Fig.~\ref{fig:map}. A raster land mask is used for map context, visualization, and route generation. Five fixed sensors are deployed: three visible-light cameras modeled using projective imaging geometry and two LiDAR units with $360^\circ$ horizontal coverage and a maximum range of $150~\mathrm{m}$. The target is a small surface vessel of size $5\times1\mathrm{~m}$, moving along Dubins-smoothed routes with speeds sampled uniformly from $[2.5,15.0]~\mathrm{km/h}$. Training, validation, and test routes are generated through water-connected regions and are followed by the simulated vessel during each episode. For evaluation, performance is reported over three spatial zones in the marina area, as shown in Fig.~\ref{fig:map}, using the same zoning rationale as in\cite{starodubov2026adaptive}.

\begin{figure*}[t]
\centering
\resizebox{\linewidth}{!}{\usetikzlibrary{
  arrows.meta,
  calc,
  fit,
  positioning,
  backgrounds,
  decorations.pathreplacing
}

\definecolor{colEnv}{RGB}{210,228,255}
\definecolor{colSen}{RGB}{255,238,200}
\definecolor{colPF} {RGB}{210,245,210}
\definecolor{colRL} {RGB}{255,215,215}
\definecolor{colBdr}{RGB}{70,70,70}

\tikzset{
  block/.style={
    rectangle, draw=colBdr, line width=0.55pt,
    rounded corners=3pt, align=center,
    font=\small\sffamily, inner sep=5pt,
    minimum height=0.9cm
  },
  arr/.style={
    -{Latex[length=2.2mm,width=1.4mm]},
    line width=0.65pt, color=black
  },
  bus/.style={line width=0.65pt, color=black},
  lbl/.style={font=\scriptsize\sffamily, fill=white, inner sep=1pt}
}

\begin{tikzpicture}

\useasboundingbox (1.0, -4.1) rectangle (15.5, 1.5);


\node[block, fill=colEnv!70, minimum width=2.0cm]
  (vessel) at (0, 0)
  {Target Vessel};

\node[block, fill=colSen!80, minimum width=1cm, minimum height=0.2cm]
  (cam) at (3.6, 0.4)
  {Camera $\times$3};

\node[block, fill=colSen!80, minimum width=1cm, minimum height=0.2cm]
  (lidar) at (3.6, -0.4)
  {LiDAR $\times$2};

\node[block, fill=colPF!80,
      minimum width=0.8cm, minimum height=2.2cm]
  (pf) at (7.0, 0.0)
  {Particle\\Filter};

\node[block, fill=colPF!40, minimum width=2.6cm, minimum height=1.5cm]
  (obs) at (10.6, 0.0)
  {Obs.\ vector $\mathbf{o}_t$\\[2pt]
   \scriptsize $[0,1]^{56}$: PF state,\\[-1pt]
   \scriptsize realized IG, coverage,\\[-1pt]
   \scriptsize sensor geometry,...};

\node[block, fill=colRL!80, minimum width=2.1cm, minimum height=2.0cm]
  (ppo) at (13.6, 0.0)
  {PPO Agent\\[5pt]
   \scriptsize Actor-Critic\\[-1pt]
   \scriptsize MLP\\[4pt]
   \scriptsize $\pi(a_t|\mathbf{o}_t)$};

\node[block, fill=colRL!35, minimum width=2.3cm]
  (rew) at (10.2, -1.8)
  {Reward $r_t$\\[2pt]
   \scriptsize $r_t = r_t^{\mathrm{IG}}$};

\node[block, fill=colSen!50, minimum width=2.6cm]
  (act) at (6.4, -2.85)
  {Action $a_t \in \{0,\ldots,4\}$\\[1pt]
   \scriptsize (all-on before PF init)};


\draw[arr] (cam.east)   -- (pf.west |- cam)
  node[lbl, midway, above] {$\mathbf{z}_\mathrm{cam}$};
\draw[arr] (lidar.east) -- (pf.west |- lidar)
  node[lbl, midway, above] {$\mathbf{z}_\mathrm{lid}$};

\draw[arr] (pf.east) -- (obs.west)
  node[lbl, midway, above] {$\hat{\mathbf{x}}_t,\sigma_t$};

\draw[arr] (obs.east) -- (ppo.west)
  node[lbl, midway, above] {$\mathbf{o}_t$};


\draw[arr]
  (pf.south) |- ($(rew.west) + (0,0.2)$)
  node[lbl, pos=0.7, below] {
  $r_t^{\mathrm{IG}}
  $};

\draw[arr]
  (rew.east) -| (ppo.south)
  node[lbl, pos=0.15, right] {$r_t$};


\draw[arr]
  (ppo.east) -- ++(0.2, 0.0) |- (act.east)
  node[lbl, pos=0.75, below] {$a_t$};


\begin{scope}[on background layer]

  \node[draw=colBdr, dashed, line width=0.45pt,
        rounded corners=4pt, fill=colSen!20,
        fit=(cam)(lidar),
        inner sep=14pt,
        label={[font=\scriptsize\sffamily\bfseries,
                text=colBdr, yshift=-12pt]above:Sensor Network}]
       (senbox) {};

  \draw[arr] (vessel.east) -- (senbox.west);
  \draw[arr] (senbox.west) -- ++(0.2,0) |- (cam.west);
  \draw[arr] (senbox.west) -- ++(0.2,0) |- (lidar.west);

  \draw[arr]
    (senbox.north) -- ++(0.0, 0.2) -| (obs.north)
    node[lbl, pos=0.25, above] {det.\ history, coverage};

  \draw[arr] (act.west) -| (senbox.south)
    node[lbl, pos=0.25, below, align=center] {selected\\sensor};

\end{scope}



\end{tikzpicture}}
\caption{Overview of the proposed IG-RLSS framework. At each decision step, the PPO agent selects one of 5 fixed sensors. The observation vector contains particle-filter-derived belief features and sensor-history features, while the particle filter maintains the vessel belief using detections and missed-detection evidence. The reward is computed from realized posterior IG.}
\label{fig:arch}
\end{figure*}
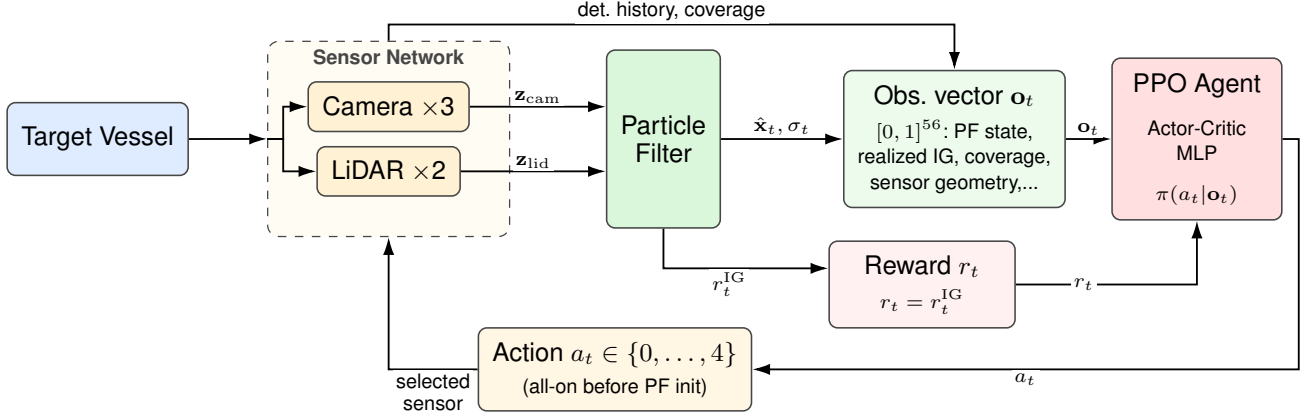

The environment is implemented as a Gymnasium-compatible episodic simulator with a standard reset--step--observe loop. The simulator time step is $\Delta t=1.0~\mathrm{s}$, and the agent makes one sensing decision every $T_d=1.0~\mathrm{s}$. During each interval, the selected sensor is activated, a detection or missed detection is generated, and the particle filter is predicted and updated. At the beginning of an episode, the vessel is placed at the start of a sampled route. Before particle-filter initialization, all sensors are on and normal single-sensor selection begins after initialization from three consecutive detection time steps. Episodes terminate when the vessel completes its route. The policy observes only belief-derived and sensor-history quantities. Ground truth is used by the simulator for measurement generation and for post hoc evaluation.

Each camera generates a two-dimensional Cartesian ground-plane measurement when the vessel projects inside the image frame and passes the visibility and stochastic detection gates. The image measurement is obtained using a calibrated pinhole projection model with base detection probability $p_d=0.95$ when the target is visible. Projected pixel coordinates are perturbed by Gaussian image noise, with additional range- and horizon-dependent uncertainty, and are then back-projected to a Cartesian measurement $\mathbf{z}=[x,y]^\top$ with an associated $2\times2$ covariance matrix. Each LiDAR unit returns a range-and-bearing measurement when the vessel lies within the vertical FOV. For both LiDAR units, the base detection probability is $p_d=0.99$; range and bearing are corrupted by Gaussian noise and converted to Cartesian coordinates for PF updating.

The vessel state is estimated using PF based on the standard sampling-importance-resampling formulation~\cite{arulampalam2002tutorial}. The PF represents the posterior over
$\mathbf{x}=[x,y,v_x,v_y]^\top$
using $N_p=500$ weighted particles. After the bootstrap phase, particles are initialized around the first accepted detection. At each subsequent time step, particles are propagated using a kinematic motion model with a turn-rate constraint of $30^\circ/\mathrm{s}$, which yields physically plausible arc-shaped particle distributions rather than the symmetric spread of an unconstrained constant-velocity Gaussian model. Camera and LiDAR detections are both supplied to the filter as Cartesian measurements with $2\times2$ covariance matrices and are evaluated using a Gaussian Cartesian likelihood. A Mahalanobis gate rejects grossly inconsistent measurements. If the active sensor produces no detection after PF initialization, a missed-detection update downweights particles lying inside that sensor's FOV according to the corresponding detection probability. Systematic resampling is triggered when
$N_{\mathrm{eff}} < 0.5N_p$.

The sensor-selection problem is formulated as a discounted Markov Decision Process
$\langle \mathcal{O}, \mathcal{A}, \mathcal{T}, \mathcal{R}, \gamma \rangle$. At each decision time step the agent receives the observation vector $\mathbf{o}_t\in \mathcal{O}$ and selects one sensor action $a_t \in \mathcal{A}= \{0,\ldots,4\}$. The single-sensor constraint is motivated by an incremental development strategy: it keeps the first learning stage interpretable and establishes a resource-constrained baseline before extending the policy to multi-sensor subset scheduling. The selected sensor remains active throughout the time step. Collected measurements are used to predict and update the PF, compute the reward $r_t\in \mathcal{R}$, and build
the next observation $\mathbf{o}_{t+1}$. Fig.~\ref{fig:arch} gives an overview of the proposed system architecture. At each decision step, the agent receives a 56-dimensional observation comprising 11 global belief features and 9 features for each of the five sensors. The global features describe PF initialization, normalized position and velocity estimates, speed, particle-cloud spread, episode time, time since the latest detection, tracking-lock status, and the realized information gain from the previous decision step. Each sensor-specific block comprises its activation state, recent detection rate and age, PF belief coverage, expected detection probability, relative range and bearing to the PF estimate, and previous-selection status. All features are derived from the PF belief, sensor geometry, and detection history, normalized to [0,1], and no ground-truth target state is provided to the policy.

The realized information gain is computed from a Gaussian approximation of the full PF state, with differential entropy evaluated from the covariance of the normalized state distribution following the standard multivariate Gaussian formulation \cite{cover2006elements}. To express the heterogeneous state components on dimensionless, comparable scales, the position and velocity components of each particle are normalized by $s_p=20~\mathrm{m}$ and $s_v=5~\mathrm{m/s}$, which are characteristic spatial and velocity scales of the tracking problem. The weighted mean $\boldsymbol{\mu}_t$ of the normalized particle cloud and its weighted $4\times4$ covariance $\boldsymbol{\Sigma}_t$ are then used to calculate the Gaussian full-state entropy of the PF belief:
\begin{equation}
H_G(b_t)
=
\frac{1}{2}
\log_2
\left[
(2\pi e)^4
\det(\boldsymbol{\Sigma}_t)
\right].
\end{equation}

At the beginning of time step $t$, the PF first performs the motion prediction, after which the entropy of the predicted particle cloud is evaluated as $H_t^-=H_G(b_t^-)$, where $b_t^-$ denotes the predicted prior belief before incorporating the current sensor evidence. After the sensing update at time step $t$, including either detection or missed-detection evidence, the posterior entropy is evaluated as $H_t^+=H_G(b_t^+)$, where $b_t^+$ denotes the resulting posterior belief. The realized full-state information gain is therefore
\begin{equation}
I_t^{\mathrm{real}}
=
H_G(b_t^-)-H_G(b_t^+).
\end{equation}
Consequently, $I_t^{\mathrm{real}}>0$ indicates a reduction in state uncertainty, $I_t^{\mathrm{real}}\approx0$ indicates little change, and $I_t^{\mathrm{real}}<0$ indicates an increase in the posterior Gaussian state uncertainty.

Finally, the reward is
\begin{equation}
r_t^{\mathrm{IG}} = 
M_t\,
\operatorname{clip}
\left(
\frac{I_t^{\mathrm{real}}}{I_{\mathrm{norm}}},
-1,1
\right),
\qquad
I_{\mathrm{norm}}=\log_2 N_p,
\end{equation}
where $I_{\mathrm{norm}}$ is the chosen information-gain normalization scale and $M_t\in\{0,1\}$ is the observability mask, equal to 0 when the target is not observable by any sensor and 1 otherwise. Thus, $I_t^{\mathrm{real}}$ denotes the realized information gain itself, whereas $r_t^{\mathrm{IG}}$ is its normalized and observability-masked reward.

The transition model $\mathcal{T}$ is induced by the simulator--particle-filter loop. After action $a_t$, the simulator propagates the vessel, applies the selected-sensor detection model, and the particle filter performs prediction, update, gating, and resampling as needed. Thus, $\mathcal{T}(\mathbf{o}_{t+1}\mid \mathbf{o}_t,a_t)$ represents the stochastic transition from the current observation and sensing action to the next belief-derived observation.

\section{Results and Analysis}
\label{sec:results}

The policy is trained with PPO~\cite{schulman2017proximal} using the Stable-Baselines3 implementation~\cite{raffin2021stable}.
The actor and critic use separate two-layer MLPs with hidden sizes $[128,128]$. The PPO hyperparameters listed in Table~\ref{tab:ppo} were initialized from standard PPO and generalized advantage estimation practice~\cite{schulman2017proximal,schulman2015high} and then adapted to the computational structure of the simulator. The learning-rate and entropy-coefficient schedules were used to encourage early exploration followed by more stable policy refinement. The discount factor  $\gamma$ was selected to emphasize long-horizon sensing decisions. Thus, the policy is encouraged to account for future information gain rather than optimizing only the immediate reward. A systematic sensitivity study is left for future work.

The agent is trained on 8000 pre-generated vessel paths, with separate validation and test splits of 1000 paths each. Checkpoints are evaluated on the validation split using deterministic rollouts with fixed path indices and fixed per-episode seeds. The selected checkpoint is the one with the lowest observable-region RMSE. Training for $2\times10^7$ environment steps required approximately 8 hours on a workstation equipped with an 12-core CPU (AMD Ryzen 9 5900X), 64 GB RAM. The 8000 paths constitute the training-route pool, whereas the computational training budget is determined primarily by the number of environment steps rather than directly by the number of stored paths.

\begin{table}[t]
  \caption{PPO Hyperparameters}
  \label{tab:ppo}
  \centering
  \small
  \begin{tabular}{lc}
    \toprule
    Hyperparameter & Value \\
    \midrule
    Learning rate $\alpha$               & linear $10^{-3}\!\rightarrow\!5\times10^{-5}$ \\
    Discount factor $\gamma$             & 0.995              \\
    GAE $\lambda$                        & 0.95               \\
    Clip range                           & 0.2                \\
    Steps per rollout $n_{\text{steps}}$ & 1000               \\
    Minibatch size                       & 2500               \\
    Entropy coefficient                  & linear $2\times10^{-2}\!\rightarrow\!2\times10^{-3}$ \\
    Value function coefficient           & 0.5                \\
    Max gradient norm                    & 0.5                \\
    Target KL                            & 0.03               \\
    Parallel environments                & 50                 \\
    Total training steps                 & $2\times10^7$       \\
    \bottomrule
  \end{tabular}
\end{table}

Four selection strategies are compared on the final test split: always-on sensing, random single-sensor selection, EIG-SS, IG-RLSS. Performance is evaluated using the RMSE between the PF-estimated and ground-truth vessel positions over non-lost time steps, together with the lost-track percentage~\cite{starodubov2026adaptive}. Always-on activates all five sensors and provides the reference tracking quality at maximum sensing cost. Random selects one sensor uniformly at each decision epoch. EIG-SS greedily selects the sensor with the largest information gain~\cite{kyriakides2021agile,starodubov2026adaptive}. IG-RLSS uses the learned policy, described in Section~\ref{sec:method}. 

\begin{figure}[t]
  \centering
  \includegraphics[width=\columnwidth]{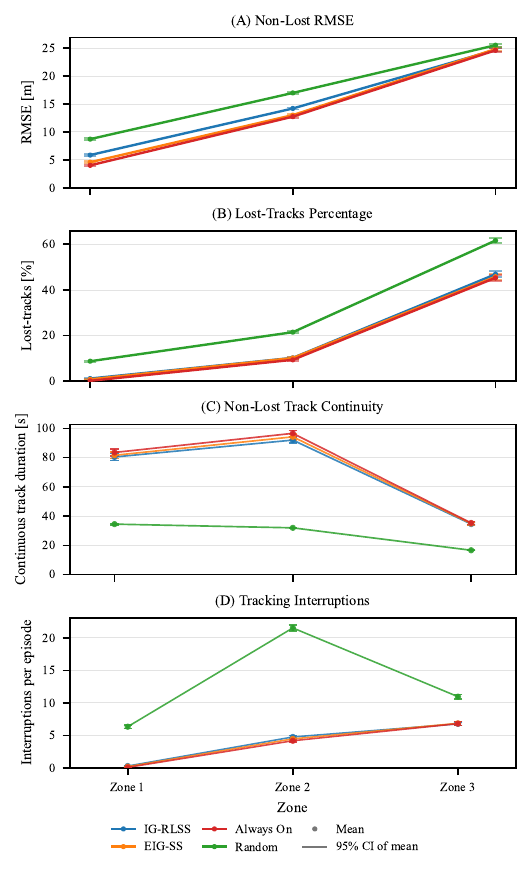}
\caption{Zone-wise tracking results on the final test split for IG-RLSS and the baseline methods. Panels report non-lost RMSE, lost-track percentage, continuous non-lost track duration, and the number of tracking interruptions per episode. Markers denote per-zone means over test episodes, and error bars denote 95\% confidence intervals of the mean.}
  \label{fig:final_test_zone_summary}
\end{figure}

Fig.~\ref{fig:final_test_zone_summary} shows that tracking difficulty increases with distance from the marina.  In Zone~1, IG-RLSS, EIG-SS, and always-on sensing achieve low non-lost RMSE of about $5~\mathrm{m}$ and almost no track loss. In Zone~2, the same methods remain close, with RMSE around $13-14~\mathrm{m}$, whereas random selection performs worse. Zone~3 is the dominant failure region: RMSE increases to about $25~\mathrm{m}$ and the lost-track percentage reaches about $45\%$ for IG-RLSS, EIG-SS, and always-on sensing, compared with more than $60\%$ for random selection. Fig.~\ref{fig:final_test_zone_summary},C and~\ref{fig:final_test_zone_summary},D show that the  IG-RLSS policy preserves track continuity close to EIG-SS and always-on sensing across all zones. In Zone 3, all non-random methods exhibit shorter continuous non-lost segments. The interruption analysis further confirms that IG-RLSS maintains temporal track stability comparable to EIG-SS and always-on sensing. The similar far-field degradation of IG-RLSS, EIG-SS, and always-on sensing indicates that Zone~3 performance appears to be mainly limited by sensing coverage rather than by the scheduling policy. To verify this interpretation, we found that in Zone 3 no sensor geometrically covered the true vessel position during 17.9\% of final-test timesteps on average, with a 95\% CI of [16.8\%, 18.9\%]. 

Because all methods were evaluated on matched test episodes, paired differences were computed as $(\Delta=\mathrm{IG\text{-}RLSS}-\mathrm{baseline})$ for each zone and metric. Holm-corrected paired $t$-tests and Wilcoxon signed-rank tests were used to assess whether the paired differences were detectably non-zero. To evaluate practical closeness rather than only detectability of differences, we additionally performed two one-sided equivalence tests for IG-RLSS versus EIG-SS and one-sided non-inferiority tests (TOST) for IG-RLSS versus always-on sensing, using pre-specified margins of 2 m for RMSE and 2\% for lost-track rate. These margins were selected as application-level tolerances for this study. The TOST results show that IG-RLSS remains equivalent to EIG-SS within these bounds in all zones, although the ordinary paired tests indicate small, around $1~\mathrm{m}$, but statistically detectable EIG-SS advantages in Zone 1 and 2. For the comparison with always-on sensing, one-sided non-inferiority tests with the same margins show that IG-RLSS remains non-inferior in all zones, while paired tests indicate that always-on still provides statistically lower error or lost-track rate in most cases. Thus IG-RLSS yields a statistically bounded approximation to EIG-SS and always-on tracking performance, while avoiding online expected IG search. 

The close performance of IG-RLSS, the information-gain-based approach, and always-on sensing is also consistent with the geometry of the considered sensor layout. In the simulated marina, the effective FOV of the fixed sensors overlap only over a limited portion of the operational area; therefore, for most target locations, the particle-filter update is dominated by either a single observing sensor or by sensors that provide limited additional independent information. Activating all sensors simultaneously therefore gives only marginal tracking benefit over selecting the most tracking-relevant sensor, while incurring the maximum sensing cost.

\begin{figure}[t]
  \centering
  \includegraphics[width=\columnwidth]{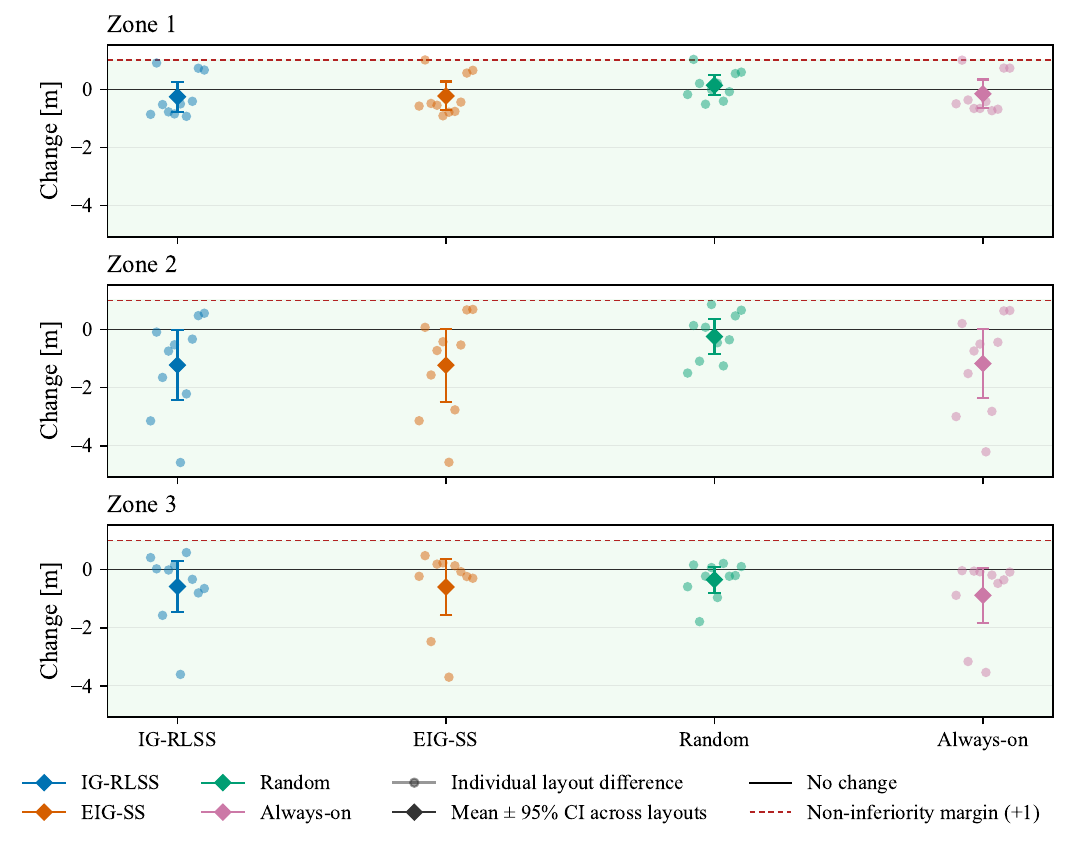}
  \caption{Zone-wise change in non-lost RMSE  for IG-RLSS and baseline methods for perturbed sensor layouts relative to the actual layout. Diamonds show mean changes with 95\% CIs whereas the red dashed line denotes the 1 meter non-inferiority margin.}
  \label{fig:sensitivity_study_for_layout}
\end{figure}

Although IG-RLSS was intentionally trained for the actual five-sensor configuration of the Ayia Napa Marina testbed rather than for arbitrary sensor deployments, we additionally evaluated the fixed policy zero-shot on 10 perturbed layouts, with camera yaw variations of $\pm15^\circ$, camera-height variations of $\pm10$~m, and LiDAR displacements within 50~m of their nominal positions. Fig.~\ref{fig:sensitivity_study_for_layout} shows the non-lost RMSE change, each dot represents one perturbed layout evaluated on matched routes/seeds. All IG-RLSS perturbations remained below the predefined $+1$~m non-inferiority margin, indicating robustness to these bounded geometric changes without retraining. Other tracking metrics remained broadly stable. Meanwhile, substantially different sensor locations, characteristics, or marina geometries may require fine-tuning or retraining. We believe, that training over randomized sensor layouts is a promising direction for developing a more layout-independent policy and will be investigated in future work. Future work will also introduce an interpretable PF-state-based myopic sensor-selection baseline to quantify whether the considered layouts can be effectively handled by simple motion- and geometry-based rules. Its comparison with IG-RLSS across layouts with different degrees of sensor overlap and decision ambiguity will help identify the conditions under which learned sequential sensor selection provides additional benefit. 

\begin{figure}[t]
  \centering
  \includegraphics[width=\columnwidth]{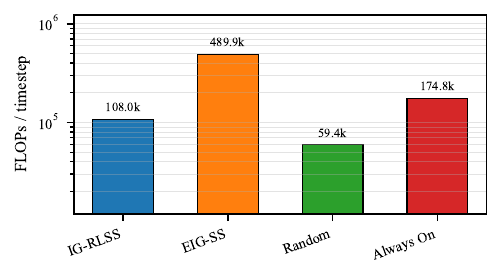}
  \caption{Estimated arithmetic cost per decision step for IG-RLSS and the baseline methods, expressed in FLOPs and including both sensor-selection cost and initialized particle-filter tracking cost.}
  \label{fig:compute_full_step_flops_runtime}
\end{figure}

Fig.~\ref{fig:compute_full_step_flops_runtime} compares the estimated full-step computational cost. For each method, the FLOP estimate was computed per decision time step as the sum of sensor-selection cost and particle-filter tracking cost. For IG-RLSS, the policy cost counts the dense actor-network operations for the observation input, including multiply-adds, bias additions, and observation normalization. For the random and always-on baselines, the cost counts only the lightweight action-selection logic. For the EIG-SS baseline, the cost includes evaluating each candidate sensor by cloning the particle filter, simulating the possible detection and missed-detection updates, estimating detection probability, and selecting the sensor with maximum expected information gain. The full-step estimate then adds the initialized particle-filter prediction, measurement weighting/update, state estimation, and resampling costs, using the measured detection and missed-update ratios from the compute benchmark. As shown in Fig.~\ref{fig:compute_full_step_flops_runtime}, the random baseline has the lowest cost, requiring only the particle-filter prediction/update/estimation workload for one active sensor. The IG-RLSS has higher cost than random selection because of the additional neural-network inference. The always-on method shows even higher cost, since the particle filter must process the measurement or missed-detection update for all five sensors. The EIG-SS is the most expensive method, mainly because it evaluates counterfactual particle-filter updates for every candidate sensor before selecting an action. It should be noted that the reported FLOP values quantify arithmetic cost only; hardware runtime, memory use, and energy consumption are left for deployment-oriented evaluation. 
For a larger sensor networks, the computational cost of both IG-RLSS and EIG-SS is expected to increase approximately linearly with the number of candidate sensors: IG-RLSS processes additional per-sensor features and policy outputs, while EIG-SS performs a separate counterfactual PF evaluation for each candidate. The exact scaling, however, depends on the sensor modalities and measurement models and should be quantified in a dedicated scalability study beyond the scope of this work.

\section{Conclusion}
\label{sec:conclusion}
This paper presents an information-guided reinforcement-learning sensor-selection framework for single-vessel tracking in a heterogeneous coastal surveillance network of fixed cameras and LiDAR units. The proposed approach couples two complementary components: a particle filter, which maintains the vessel belief from heterogeneous measurements, and a PPO-based policy, which learns to select the sensing asset most useful for maintaining that belief. Realized posterior entropy reduction is used as an information-gain signal, so the policy is trained from belief-state uncertainty and sensor-use utility rather than from an ground-truth tracking-error penalty.

On the held-out test set, the proposed IG-RLSS framework achieved tracking performance comparable to the explicit EIG-SS baseline and close to always-on sensing in the main operational zones, while activating only one sensor per decision time step. Unlike EIG-SS, the learned policy avoids repeated online counterfactual particle-filter entropy searches, providing a lower-complexity approximation to information-driven sensor management.

The present study assumes fixed camera detection characteristics and does not explicitly model illumination-dependent perception, although visible-light detection performance can vary substantially between daytime and low-light/night conditions. Evaluating condition-specific visual detectors is beyond the scope of this work. The proposed framework is initially designed to be modular with respect to the sensing front end, allowing such detectors and other sensing modalities to be incorporated in future studies without changing the underlying sensor-selection principle.

This study establishes a belief-based single-sensor scheduling baseline and a georeferenced simulation framework for resource-aware maritime surveillance. Future work will also extend IG-RLSS to energy-aware multi-sensor subset scheduling across alternative sensor geometries and validate it under real maritime conditions.

\section*{Acknowledgment}

This work was supported by the LORELEI-X project (Grant Agreement No. 101159489), funded by the EU under the Horizon Programme and by MDigi-I project ``Maritime Digitalization Research Infrastructure'' (STRATEGIC INFRASTRUCTURES/1222/0113), implemented under the Cohesion Policy Programme ``THALIA 2021--2027'' and co-funded by the Republic of Cyprus and the European Regional Development Fund (ERDF).

CMMI was established as a “Center of Excellence” in Marine and Maritime Research, Technology Development \& Innovation and has received funding from the EU Horizon 2020 program under grant agreement No. 857586 and matching funding from the Government of the Republic of Cyprus.

\bibliographystyle{IEEEtran}
\bibliography{main_reference}

@article{starodubov2026adaptive,   
    title={Adaptive Entropy-Driven Sensor Selection in a Camera-LiDAR Particle Filter for Single-Vessel Tracking},   
    author={Starodubov, Andrei and Prabowo, Yaqub Aris and Hadjipieris, Andreas and Kyriakides, Ioannis and Galeazzi, Roberto},   
    journal={arXiv preprint arXiv:2603.08457},   
    year={2026} 
}

@inproceedings{kyriakides2021agile,   
    title={Agile Target Tracking Based on Greedy Information Gain},   
    author={Kyriakides, Ioannis},   
    booktitle={2021 IEEE 12th Annual Information Technology, Electronics and Mobile Communication Conference (IEMCON)},   
    pages={0204--0208},   
    year={2021},   
    organization={IEEE} 
}

@inproceedings{chang2003vessel,
  title={Vessel identification and monitoring systems for maritime security},
  author={Chang, Shwu-Jing},
  booktitle={IEEE 37th Annual 2003 International Carnahan Conference on Security Technology, 2003. Proceedings.},
  pages={66--70},
  year={2003},
  organization={IEEE}
}

@article{lyu2022sea,
  title={Sea-surface object detection based on electro-optical sensors: A review},
  author={Lyu, Hongguang and Shao, Zeyuan and Cheng, Tao and Yin, Yong and Gao, Xiaowei},
  journal={IEEE Intelligent Transportation Systems Magazine},
  volume={15},
  number={2},
  pages={190--216},
  year={2022},
  publisher={IEEE}
}

@article{schulman2017proximal,
  title={Proximal policy optimization algorithms},
  author={Schulman, John and Wolski, Filip and Dhariwal, Prafulla and Radford, Alec and Klimov, Oleg},
  journal={arXiv preprint arXiv:1707.06347},
  year={2017}
}

@article{raffin2021stable,
  title={Stable-baselines3: Reliable reinforcement learning implementations},
  author={Raffin, Antonin and Hill, Ashley and Gleave, Adam and Kanervisto, Anssi and Ernestus, Maximilian and Dormann, Noah},
  journal={Journal of machine learning research},
  volume={22},
  number={268},
  pages={1--8},
  year={2021}
}

@article{schulman2015high,
  title={High-dimensional continuous control using generalized advantage estimation},
  author={Schulman, John and Moritz, Philipp and Levine, Sergey and Jordan, Michael and Abbeel, Pieter},
  journal={arXiv preprint arXiv:1506.02438},
  year={2015}
}

@article{zheng2023novel,
author = {Zheng, L. and Liu, M. and Zhang, S. and Lan, J.},
title = {A Novel Sensor Scheduling Algorithm Based on Deep Reinforcement Learning for Bearing-Only Target Tracking in {UWSNs}},
journal = {IEEE/CAA Journal of Automatica Sinica},
volume = {10},
number = {4},
pages = {1077--1079},
year = {2023},
doi = {10.1109/JAS.2023.123159}
}

@inproceedings{hoffmann2020sensorpath,
author = {Hoffmann, F. and Charlish, A. and Ritchie, M. and Griffiths, H.},
title = {Sensor Path Planning Using Reinforcement Learning},
booktitle = {Proceedings of the IEEE 23rd International Conference on Information Fusion},
pages = {1--8},
year = {2020},
doi = {10.23919/FUSION45008.2020.9190242}
}

@article{qu2022improved,
author = {Qu, Z. and Zhao, X. and Xu, H. and Tang, H. and Wang, J. and Li, B.},
title = {An Improved {Q-Learning-Based} Sensor-Scheduling Algorithm for Multi-Target Tracking},
journal = {Sensors},
volume = {22},
number = {18},
pages = {6972},
year = {2022},
doi = {10.3390/s22186972}
}

@misc{jeong2020learning,
author = {Jeong, Heejin and Hassani, Hamed and Morari, Manfred and Lee, Daniel D. and Pappas, George J.},
title = {Learning to Track Dynamic Targets in Partially Known Environments},
year = {2020},
archivePrefix = {arXiv},
eprint = {2006.10190},
primaryClass = {cs.RO},
}

@inproceedings{yang2023policy,
  title = 	 {Policy Learning for Active Target Tracking over Continuous $SE(3)$ Trajectories},
  author =       {Yang, Pengzhi and Koga, Shumon and Asgharivaskasi, Arash and Atanasov, Nikolay},
  booktitle = 	 {Proceedings of The 5th Annual Learning for Dynamics and Control Conference},
  pages = 	 {64--75},
  year = 	 {2023},
  editor = 	 {Matni, Nikolai and Morari, Manfred and Pappas, George J.},
  volume = 	 {211},
  series = 	 {Proceedings of Machine Learning Research},
  month = 	 {15--16 Jun},
  publisher =    {PMLR}
}

@article{ma2020dpfrl,
  title={Discriminative Particle Filter Reinforcement Learning for Complex Partial Observations},
  author={Xiao Ma and Peter Karkus and David Hsu and Wee Sun Lee and Nan Ye},
  journal={ArXiv},
  year={2020},
  volume={abs/2002.09884},
}

@inproceedings{jonschkowski2018differentiable,
author = {Jonschkowski, Rico and Rastogi, Divyam and Brock, Oliver},
title = {Differentiable Particle Filters: End-to-End Learning with Algorithmic Priors},
booktitle = {Proceedings of Robotics: Science and Systems},
year = {2018},
address = {Pittsburgh, Pennsylvania},
month = jun,
doi = {10.15607/RSS.2018.XIV.001},
archivePrefix = {arXiv},
eprint = {1805.11122},
primaryClass = {cs.LG},
}

@inproceedings{fischer2020ipft,
author = {Fischer, Johannes and Tas, {"O}mer Sahin},
title = {Information Particle Filter Tree: An Online Algorithm for {POMDP}s with Belief-Based Rewards on Continuous Domains},
booktitle = {Proceedings of the 37th International Conference on Machine Learning},
pages = {3177--3187},
year = {2020},
editor = {Daum{'e} III, Hal and Singh, Aarti},
volume = {119},
series = {Proceedings of Machine Learning Research},
publisher = {PMLR},
}

@article{arulampalam2002tutorial,
author = {Arulampalam, M. S. and Maskell, S. and Gordon, N. and Clapp, T.},
title = {A Tutorial on Particle Filters for Online Nonlinear/Non-{Gaussian} Bayesian Tracking},
journal = {IEEE Transactions on Signal Processing},
volume = {50},
number = {2},
pages = {174--188},
year = {2002},
month = feb,
doi = {10.1109/78.978374}
}

@book{cover2006elements,
  author    = {Thomas M. Cover and Joy A. Thomas},
  title     = {Elements of Information Theory},
  edition   = {2},
  publisher = {John Wiley \& Sons},
  address   = {Hoboken, NJ, USA},
  year      = {2006},
  doi       = {10.1002/047174882X}
}

\end{document}